**Modular, Multi-Robot Integration of Laboratories: An Autonomous Solid-State Workflow for Powder X-Ray Diffraction**

Amy. M. Lunt,[1,2] Hatem Fakhruldeen,[1] Gabriella Pizzuto,[1] Louis Longley,[1] Alexander White,[1] Nicola Rankin,[1,2] Rob Clowes,[1] Ben Alston,[1,2] Lucia Gigli,[3] Graeme M. Day,[3] Andrew I. Cooper[1,2]* and Sam. Y. Chong[1,2]*

[1] Department of Chemistry and Materials Innovation Factory, University of Liverpool, L7 3NY, UK. E-mail: aicooper@liverpool.ac.uk, schong@liverpool.ac.uk

[2] Leverhulme Research Centre for Functional Materials Design, University of Liverpool, Liverpool L7 3NY, UK

[3] Computational Systems Chemistry, School of Chemistry, University of Southampton, SO17 1BJ, UK

Automation can transform productivity in research activities that use liquid handling, such as organic synthesis, but it has made less impact in materials laboratories, which require sample preparation steps and a range of solid-state characterization techniques. For example, powder X-ray diffraction (PXRD) is a key method in materials and pharmaceutical chemistry, but its end-to-end automation is challenging because it involves solid powder handling and sample processing. Here we present a fully autonomous solid-state workflow for PXRD experiments that can match or even surpass manual data quality. The workflow involves 12 steps performed by a team of three multipurpose robots, illustrating the power of flexible, modular automation to integrate complex, multitask laboratories.

Robots can carry out a range of repetitive and iterative laboratory tasks, particularly those involving liquid handling (*1*), such as for organic synthesis (*2–5*), but many experiments remain hard to automate. One example is powder X-ray diffraction (PXRD), which is a central tool for characterizing the structure of ordered solids (*6*), including functional materials (*7,8*) and pharmaceutical polymorphs (*9*). PXRD can be easier to implement than methods that require growing and harvesting a single diffractable crystal (*10*), and it provides important information about structure and purity (*11–13*). PXRD is used for the rapid identification of crystal forms and for detecting the existence of polymorphs (*14–16*); this is valuable in both materials research (*7,8,17,18*) and in pharmaceutical chemistry (*19,20*). Indeed, pharmaceutically active molecules must undergo exhaustive and expensive screening experiments to fully understand their crystal form landscapes before they can be approved for clinical trials (*21–23*).

Previously, high-throughput crystallization screens have used robots and other automated platforms (*16,17,24–26*) to accelerate the discovery of materials such as pharmaceuticals (*21–26*), porous organic cages (*27*) and photovoltaics (*28*), but these workflows tend to be only partially automated. For closed-loop autonomous workflows, we must automate and connect all stages in the PXRD experiment. This starts with crystal growth and is followed by sample preparation, often by mechanical grinding to reduce the crystal size to allow better orientational averaging. The resulting powders are then transferred into a sample holder, such as a multi-well plate, followed by loading the PXRD instrument and data collection. At present, this sample preparation and transfer is typically done by hand, even for 'high throughput', automated workflows (*17*), and this is laborious. Likewise, there is a plethora of other materials workflows where similar solid-handling operations are required, such as preparing samples for conductivity analysis or for microscopy.

Here, we present a fully autonomous, modular robotic workflow that prepares crystalline materials and then collects their powder diffraction data. This modular approach integrates three separate robotic platforms—a liquid handling platform for the crystallization stage (Chemspeed FLEX LIQUIDOSE, Figure 1, step 1 & Supporting Information, Figures S1 & S2), a mobile manipulator for sample transport and equipment manipulation (KUKA KMR iiwa; Fig. 1, steps 1, 2, 8 & 9–12), and a dual-arm robot for sample preparation (ABB YuMi; Fig. 1, steps 3–7). The workflow uses a standard powder X-ray diffractometer, which is used by the mobile manipulator in an anthropomorphic way without any substantial modification. These heterogenous robotic and automation platforms work together synchronously to achieve the multiple steps in the workflow (Figure 1), orchestrated by our system architecture, ARChemist (Supporting Information, Figure S3) (*29*). We exemplify this approach with two different organic compounds and show that PXRD data collected by the autonomous robotic workflow are of comparable quality, or in some cases better, to data collected for samples prepared by hand. Hence, these data are suitable for identifying compounds and distinguishing between their polymorphs. We also demonstrate matching crystalline powders against sets of putative polymorphs generated by crystal structure prediction (CSP) methods, which is a key step in integrating predictive computational methods into closed-loop materials discovery.

**Description of the modular robotic workflow**

The workflow comprises three robots and 12 steps, as outlined in Figure 1 and the videos in the Supporting Information (Videos 1 & 2). First, crystals are grown using a Chemspeed platform (Supporting Information, Figure S1,2), whereby the material of interest is dispensed in a variety of solvents or solvent mixtures and these solvents are allowed to evaporate, thus growing crystals (Step 0, not shown). Quite often this leads to large crystals that adhere strongly to the sides of the sample vials (Supporting Information, Figure S4) and these must therefore be reduced in size and removed from the vial prior to diffraction, as described below. In Step 1 (Fig. 1), a rack of eight crystal samples is collected from the Chemspeed platform using a mobile KUKA robot (*30*); the Chemspeed platform was modified with an automated vertical sash door to facilitate this. Each of the eight vials is capped with a sample lid sealed with an adhesive Kapton polymer film that will ultimately receive the ground crystalline powder (Supporting Information, Figures S4,5). In Step 2, the mobile KUKA manipulator delivers the rack of eight samples to the preparation station, which involves a dual-arm ABB YuMi robot. In Step 3, the dual-arm YuMi robot transfers the eight samples to grinding station 1, where they are reduced in size using mechanical attrition by magnetic stirring with preloaded Teflon stir bars.

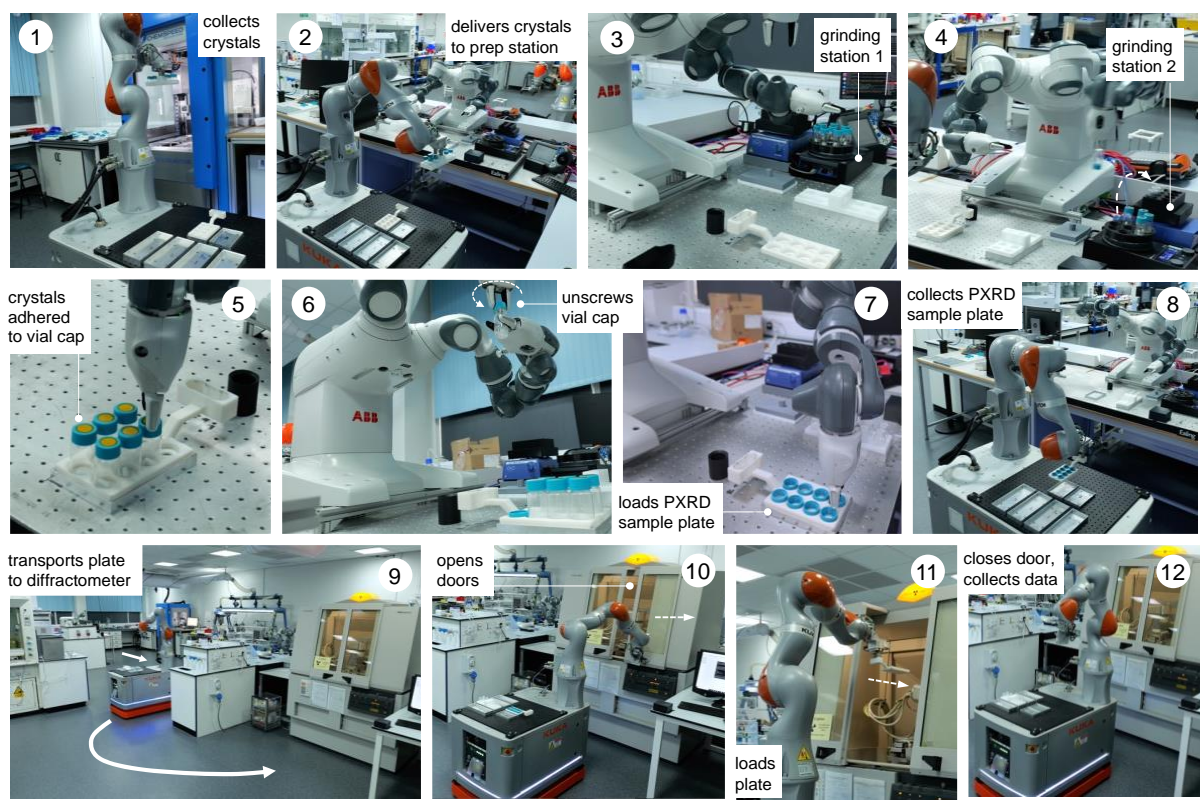

**Fig. 1**. **Multi-robot workflow for autonomous crystal growth, sample preparation and powder X-ray diffraction.** It comprises 12 steps and integrates three separate robots, orchestrated by our autonomous robotic chemist system architecture, ARChemist.

In Step 4, the YuMi robot inverts the eight samples and transfers them to grinding station 2, where they are agitated using a shaker plate to reduce the particle size further and to transfer the sample onto the adhesive Kapton polymer film in the cap of each vial. In Step 5, the YuMi robot inverts the samples again and transfers them to the X-ray diffraction plate; at this point, the sample is adhered to the Kapton film in the vial cap (Supporting Information, Figure S4) and any excess sample and the Teflon stir bar falls back into the vial. In Step 6, the YuMi robot unscrews each vial cap, inverts it, and places it back into the PXRD plate (Step 7, Supporting Information, Figure S6), which is then collected by the KUKA KMR iiwa robot (Step 8) and transported to the diffractometer (Step 9). In Step 10, the KUKA robot opens the sliding doors of the diffractometer, loads the plate into the instrument (Step 11), and then closes the doors and X-ray data is collected for the eight samples (Step 12). The sample rack can then be retrieved from the PXRD instrument by the KUKA robot and another rack of eight samples processed and analysed, as required. A full loop of sample preparation, transport, and data collection takes around nine hours for a rack of eight samples with the data acquisition settings used here, although this time depends on the scan parameters; here, 1 hour for sample processing plus eight 1-hour PXRD scans. The overall workflow is detailed by the videos in the Supporting Information (Videos 1 & 2) and associated process flow diagrams (Figures S7 & S8). As shown in Figure 2, this autonomous PXRD workflow is part of a larger laboratory that contains several other workflows—for example, our autonomous photocatalysis workflow (*30*) is in the same area as the X-ray diffractometer (labelled **C** in Figure 2).

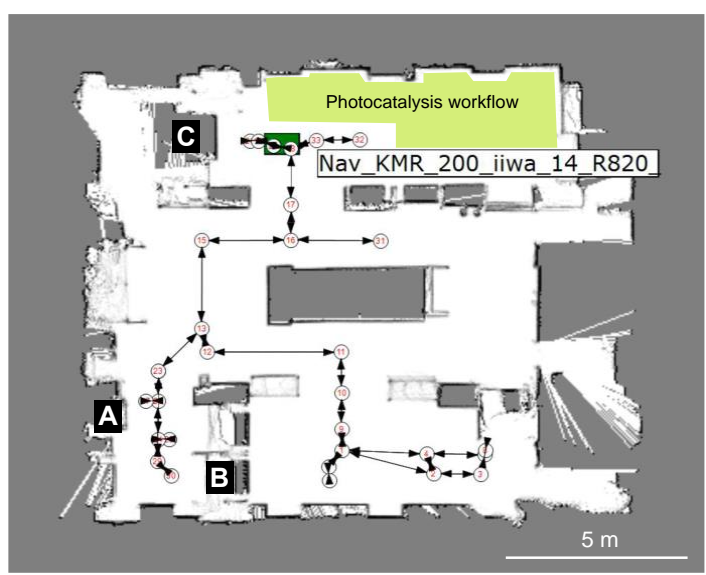

**Fig. 2**. **Heterogenous modular integration of laboratories using multiple robots.** Image from the KUKA Sunrise software showing the robot-generated map of the lab with numbered nodes and edges that correspond to taught paths and way points. For the experiments described here, the key modules are the Chemspeed liquid handling platform (**A**), the YuMi dual-arm sample preparation station (**B**) and the powder X-ray diffractometer (**C**). The location of the KUKA KMR iiwa robot is shown by the green rectangle, in this case approaching the X-ray diffractometer, **C**. The location of our photocatalysis workflow (not discussed here) is shaded in yellow (*30*).

The layout of the workflow is, to some extent, arbitrary: for example, the location of the X-ray diffractometer is fixed by proximity to its cooling water supply, and there was insufficient space to locate the Chemspeed FLEX LIQUIDOSE robot (**A**, Figure 2) or the ABB YuMi preparation station (**B**) adjacent to the diffractometer. This does not matter under a modular paradigm that uses mobile manipulators to integrate stations because the transport time between the stations is a small fraction of the overall workflow cycle time compared to the slow steps, which in this case are solvent evaporation for crystallization and PXRD data acquisition. The approach is therefore inherently scalable: for example, one can envisage coupling two workflows together, whereby the most crystalline materials are selected for testing as photocatalysts (*30*) and those powders transferred automatically into that workflow, which is housed in the same laboratory (yellow shaded area in Figure 2). Also, because we use collaborative robots, or 'cobots', the laboratory space can be shared with human researchers, and we do this daily.

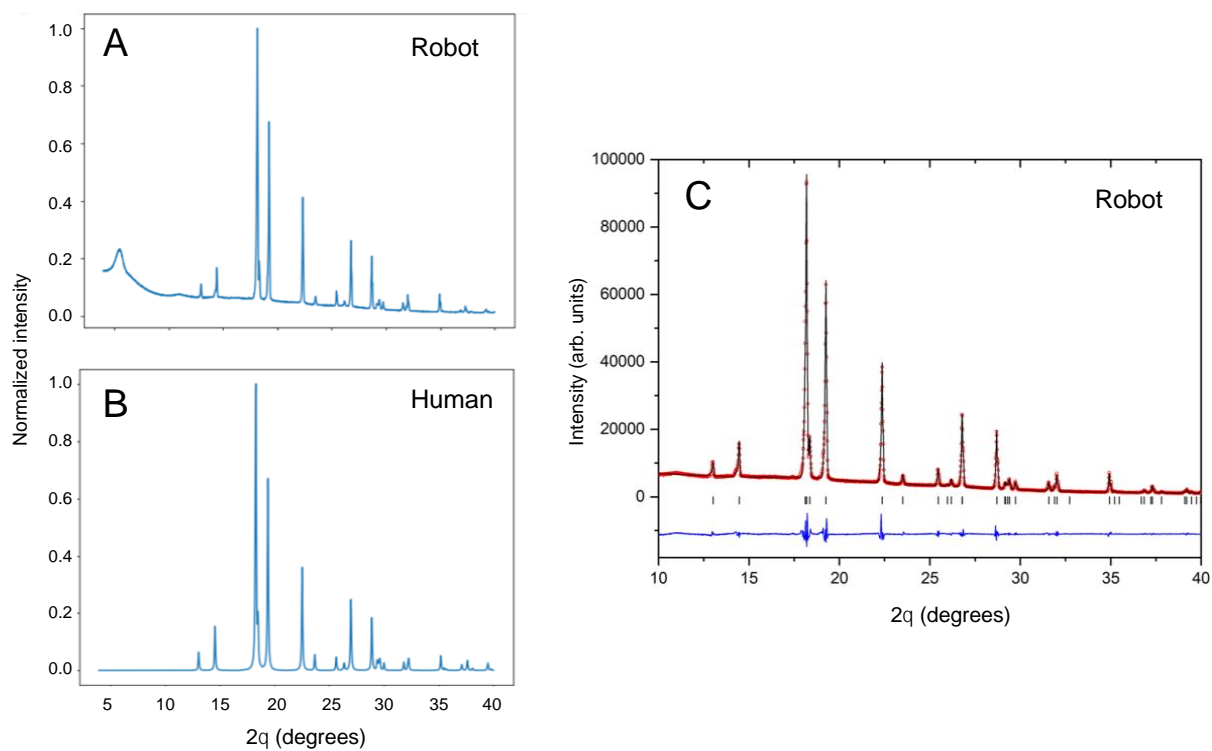

**Fig. 3**. **Comparison of powder X-ray diffraction patterns for samples prepared by robots and by humans.** (**A**) Data collected using the autonomous robotic workflow (crystallization, grinding, and sample mounting) for the alpha polymorph of benzimidazole and (**B**) data collected by conventional manual methods including grinding using a pestle and mortar. (**C**) Final observed (red circles), calculated (black line) and difference (blue) profiles from Le Bail refinement of the PXRD data from the robot-prepared benzimidazole sample. Tick marks indicate reflection positions.

**Automated data collection for benzimidazole**

Benzimidazole is an organic heterocyclic compound that typically exists as its alpha polymorph in the form of solid white crystals. Benzimidazole derivatives are used in pharmaceuticals such as antacids, antiparasitic drugs and opioids (*31–33*). A stock solution of benzimidazole in methanol (0.1 g/mL) was used for the automated experiment. The input station in the Chemspeed FLEX LIQUIDOSE platform was loaded with a rack of eight sample vials (20 mL vial volume) preloaded with magnetic Teflon stir bars and capped with Kapton film vial lids, as described above. The stock solutions were dispensed into the eight sample vials and left to evaporate inside the Chemspeed platform. Once the solid samples were dry, the vials were capped by the Chemspeed platform. Like many organic compounds, benzimidazole often crystallizes as large, blocky crystals that are hard to recover and unsuitable for PXRD analysis without further preparation, as shown in the Supporting Information (Fig. S4).

Next, the mobile KUKA manipulator collects the samples from the Chemspeed platform and delivers them to the preparation station for processing, as described above, followed by automated PXRD analysis. To compare our autonomous method with the traditional manual approach, 2 mL of the stock solution was also dispensed into a sample vial and left to evaporate, whereafter the solid was recovered and ground by hand using a mortar and pestle. This sample was mounted, also by hand, in a 96-well aluminium plate prior to data collection. Data collected under the same scan conditions for robot-prepared and human-prepared benzimidazole crystals are compared in Figure 3A,B.

The broad peak at around $2\theta = 6°$ in the robot-collected data results from the adhesive Kapton tape on which the robot-prepared samples are loaded. Apart from this artefact, the two datasets are comparable in terms of peak width, peak positions, and relative peak intensities. The PXRD pattern generated by the autonomous robotic workflow shows good signal to noise and is of sufficient quality to extract structural information (Figure 3C). The unit cell parameters that were determined (a = 13.6000(5) Å; b = 6.8564(2) Å; c = 6.9905(2) Å) confirmed the formation of the alpha polymorph by comparison to reference structures in the CSD database (Supporting Information, Table S1). The analysis showed that the sample prepared by the automated process had better homogeneity in terms of crystallite size, resulting in more consistent peak profiles that could be better modelled. This enabled more precise lattice parameters with smaller standard uncertainties to be determined than from the conventional, manually prepared sample (Supporting Information, Table S1).

**Automated data collection for polymorphic ROY**

The archetypal example a polymorphic organic molecule is 5-methyl-2- [(2-nitrophenyl) amino]-3-thiophenecarbonitrile, commonly referred to as ROY (*34–38*). It is an intermediate in the synthesis of the antipsychotic drug olanzapine and it is named ROY because of its red, orange, and yellow polymorphs. There have been many crystallization studies conducted for ROY, and various polymorphs have been observed, which often occur concomitantly with each other as mixtures. Here, samples of ROY were prepared from solid 5-methyl-2-[(2-nitrophenyl) amino]-3-thiophenecarbonitrile by dissolving it in acetone with the addition of different percentage volumes of water as an antisolvent (Supporting Information, Table S2). In this case, the sample solutions were prepared manually because of the slow evaporation time for water, which would require the Chemspeed robot to be idle over long periods of time, although this step could have been automated as for benzimidazole, above. After a dry, solid product had formed, the samples were loaded into the input station for the workflow for processing and PXRD analysis, as before (Figure 4). This experiment produced two different polymorphs of ROY, and the data obtained by the robot was of sufficient quality to identify them, even when appearing as a mixture (*e.g.*, sample 4 in Figure C,D). As for benzimidazole, unit cell parameters extracted from the PXRD data for sample 1 confirmed the effective homogenization of the powder and its phase purity (Supporting Information, Figure S14, Table S3); again, the robot-prepared data was superior to the manually produced samples, resulting in improved fit statistics and uncertainties. The data quality across the batch of eight samples can be seen in Figure 4D.

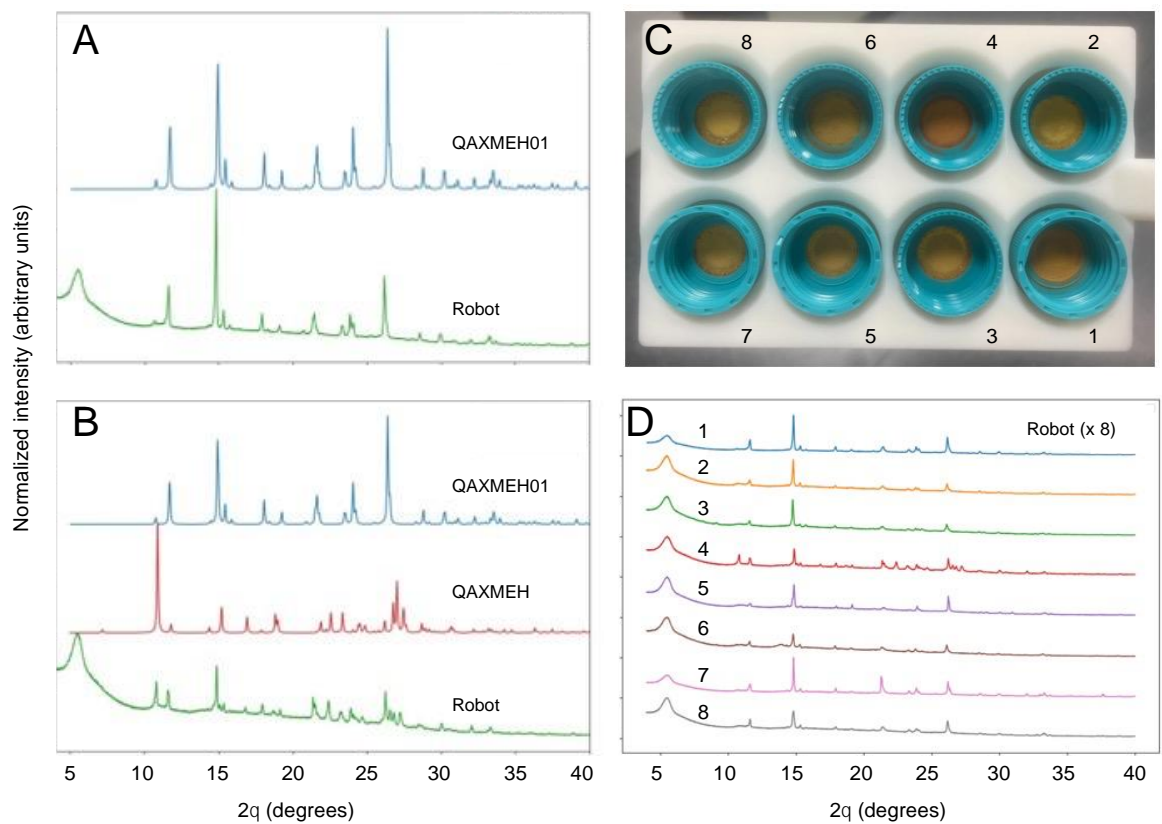

**Fig. 4**. **PXRD patterns collected autonomously for ROY polymorphs.** (**A**) Comparison of the diffraction pattern for ROY processed using the robotic workflow (sample 1 in Fig. 4C,D), as compared with the simulated PXRD pattern for the published monoclinic yellow (Y) polymorph (CSD reference code QAXMEH); (**B**) PXRD patterns for ROY processed using the autonomous robotic workflow (sample 4 in Fig. 4C,D), as compared with the simulated patterns for two published forms: the monoclinic Y polymorph (QAXMEH) and the monoclinic orange needle (ON) polymorph (QAXMEH01), suggesting that a phase mixture is formed under these conditions; (**C**) Photograph of ROY processed using the robotic workflow at various concentrations and solvent ratios, see Supplementary Information, Table S1, for crystallization conditions); (**D**) Diffraction patterns for the ROY samples shown in Fig. 4C.

## Polymorph matching against crystal structure prediction datasets

Crystal structure prediction (CSP) is valuable for anticipating polymorphism of active pharmaceutical ingredients (*39*) and in guiding the discovery of molecular crystals with targeted properties (*40*). We therefore assessed the possibility of comparing the powder X-ray diffraction patterns generated by the automated robotic workflow against structure sets of putative crystal structures generated by CSP. From sample 1 for benzimidazole and ROY, we found that comparison of the experimental data against the low energy CSP structures using the variable-cell experimental powder difference (VC-xPWDF) method (*41*) identifies the predicted crystal structure as the observed polymorph; that is, alpha benzimidazole (Figure 5A) has the most similar simulated powder X-ray diffraction pattern. Likewise, sample 1 of ROY can be identified as matching the predicted low-energy Y polymorph most closely (Figure 5B). This demonstrates the possibility of identifying newly discovered crystal structures in an automated manner by comparison against pre-computed libraries of predicted crystal structures, which would create an important feedback mechanism between computational screening of materials and automated crystallization in the laboratory.

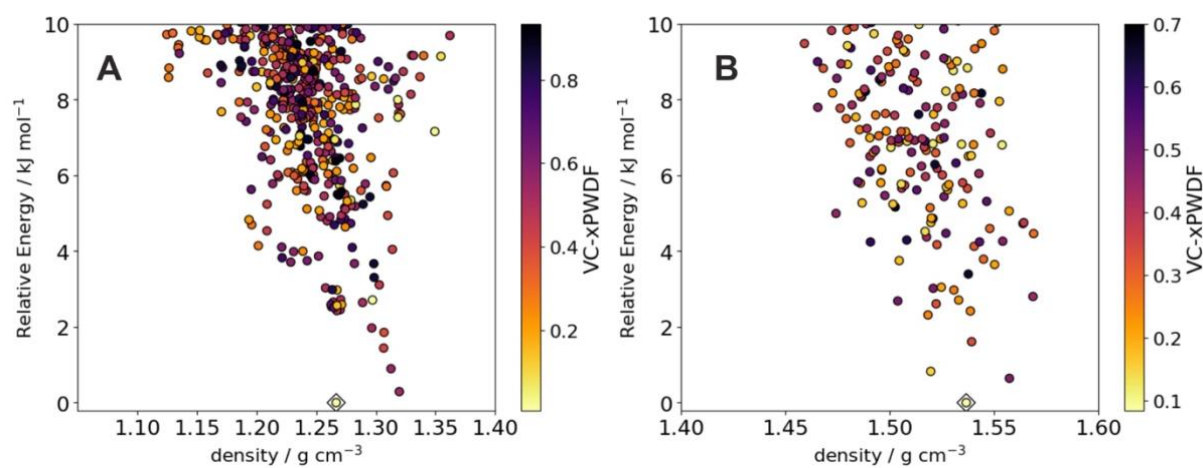

**Fig. 5**. **Matching robotic PXRD data with computationally predicted crystal structures.** Energy-density distribution of low-energy crystal structure prediction (CSP) structures of (**A**) benzimidazole and (**B**) ROY. Each point corresponds to a distinct predicted crystal structure; points are colored by dissimilarity of their simulated powder X-ray diffraction patterns compared to the pattern collected from the robot workflow. For ROY, we show results using pattern 1 from Figure 4. The CSP structures corresponding to the alpha polymorph of benzimidazole and the Y polymorph of ROY are indicated with diamonds; in both cases, these correspond to the lowest dissimilarity (greatest similarity) to the experimental data.

**Outlook**

This work illustrates the potential of modular and flexible robots (*42*) to accelerate PXRD experiments, thus integrating a key materials characterization method into a 'self-driving laboratory' (*43*). While the videos presented here show a single cycle (one rack of eight samples), it should be possible in the future to operate this workflow 24/7 over extended periods in a closed-loop way, as demonstrated in our earlier implementation of mobile robots for photocatalyst optimization (*30*). Some method improvements will be necessary for fully autonomous matching with predicted crystal structures, which are sensitive to sample preparation and polymorphic purity. For example, for the three ROY samples that we identified by eye as pure polymorph Y, PXRD matching was able to identify the correct CSP structure. However, the matching algorithm identifies incorrect structures when presented with PXRD of samples of a polymorphic mixture, such as sample 4 in Figure 4D. Even then, however, the CSP structure corresponding to polymorph Y was identified among the best matches to the experimental PXRD (9$^{th}$ out of 264 CSP structures), albeit not as the best match. Likewise, some samples of benzimidazole that we tested had different relative peak intensities their PXRD pattern, probably because of a non-uniform distribution of crystallite orientations, which led to less reliable matching to the CSP structure that corresponds to alpha benzimidazole (Supplementary Information, Table S4). These challenges should be addressable via improvements to the sample preparation methods and the PXRD matching algorithms.

The introduction of sample processing and the use of three separate robots rather than one makes this solid-state workflow significantly more complex than our earlier photocatalysis study (*30*), which comprised a single, easily automated measurement (gas chromatography) and no sample processing. Indeed, this workflow ranks among the most complex autonomous systems reported for chemistry to date (*2–5,42–45*), and its modular nature offers unique scope for expansion and diversification. An automated platform of comparable complexity is AMANDA (*44*), which is being developed for photovoltaics research; another impressive example is the AutoBASS platform that can assemble up to 64 CR2023 battery cells (*45*). However, both of those platforms involve large, custom-built integrated robotic systems. By contrast, our workflow uses commercially available, 'off the shelf' robots and other common laboratory hardware with little or no modification. By using a mobile manipulator to integrate the various stations, the workflow can be arranged in almost any configuration, and it is easily extended by adding other stations, subject only to available laboratory space (Figure 2). As such, we see the general concept of integrating stationary and mobile robots using a core software architecture (*29*) as a powerful strategy for automating a range of research activities beyond diffraction experiments. This will also allow us to guide autonomous closed-loop robotic experiments using computational predictions and artificial intelligence (*44*). To give just one example, it should be possible to use this autonomous PXRD workflow to identify crystallization conditions that produce polymorphs that are predicted to have certain desirable functional properties (*40*), and then to automatically take those materials forward for property testing.

## Acknowledgments

We thank Prof. K. Thurow (University of Rostock) for advice and for inspiring us to work with mobile robots.

## Funding

This project has received funding from the European Research Council under the European Union's Horizon 2020 research and innovation program (grant agreement no. 856405). The authors also received funding from the Engineering and Physical Sciences Research Council (EPSRC) (EP/V026887/1, EP/T031263/1) and the Leverhulme Trust via the Leverhulme Research Centre for Functional Materials Design. AIC thanks the Royal Society for a Research Professorship (RSRP\S2\232003).

## Author contributions

S.Y.C. and A.I.C. conceived the project. A. M. L. led the experimental work, the design of the dual-arm robotic processing station, and the overall construction of the workflow. L.L. and A.W. also contributed to the design of the processing station. N.R. assisted with design and 3D printing of components in the workflow. R.C. assisted with the PXRD and Chemspeed integrations. B.M.A. was involved in the overall workflow design. H.F. and G.P. contributed to the robotics aspects. H.F. led the design and implementation of the ARChemist software suite. A.M.L. and S.Y. performed the LeBail fits for the PXRD data. G.M.D. and L.G. carried out the crystal structure prediction (CSP) work and structural matches with CSP data. All authors contributed to the writing of the paper.

## Competing interests

The authors declare no competing interests.

## Data and materials availability

All data needed to evaluate the conclusions in the paper are present in the paper or the supplementary materials and the repository links therein.

# Supporting Information

**Modular, Multi-Robot Integration of Laboratories: An Autonomous Solid-State Workflow for Powder X-Ray Diffraction**

Amy. M. Lunt,[1,2] Hatem Fakhruldeen,[1] Gabriella Pizzuto,[1] Louis Longley,[1] Alexander White,[1] Nicola Rankin,[1,2] Rob Clowes,[1] Ben Alston,[1,2] Lucia Gigli,[3] Graeme M. Day,[3] Andrew I. Cooper[1,2]* and Sam. Y. Chong[1,2]*

[1] Department of Chemistry and Materials Innovation Factory, University of Liverpool, L7 3NY, UK. E-mail: aicooper@liverpool.ac.uk, schong@liverpool.ac.uk

[2] Leverhulme Research Centre for Functional Materials Design, University of Liverpool, Liverpool L7 3NY, UK

[3] Computational Systems Chemistry, School of Chemistry, University of Southampton, SO17 1BJ, UK

**Contents**

# 1. Supporting Videos

**Video 1 (6 min 16 sec, recorded at 4K resolution)**: Captioned video overview of the 12 stages of the autonomous PXRD workflow described in Fig. 1 in the main paper, along with the associated text. Note that the video is accelerated by varying degrees at different stages of the workflow, but the average acceleration is approximately x 10. An unaccelerated version of this Video without captions (55 min 29 sec) is included below. Benzimidazole was the test substance used in these experiments.

YouTube link: https://youtu.be/8rnaoTF-VMk

Original video file (downloadable):

https://www.dropbox.com/s/riq3jktroa74dbs/PXRD_6%27_16%22_captions_4K.mp4?dl=0

The key steps in the workflow can be found at the following points in **Video 1** (*note these are also marked as clickable **time stamps** in the online video*).

| | |
|---|---|
| **0 min 0 sec** | **Step 1:** KUKA robot collects a rack of 8 crystalline samples from the Chemspeed platform. |
| **0 min 24 sec** | **Step 2:** Mobile KUKA manipulator delivers the rack of 8 samples to the preparation station, which involves a dual-arm ABB YuMi robot. |
| **1 min 6 sec** | **Step 3:** Dual-arm YuMi robot transfers the 8 samples to grinding station 1, where they are reduced in size using mechanical attrition (magnetic stirring). |
| **1 min 40 sec** | **Step 4:** The YuMi robot inverts the 8 samples and transfers them to grinding station 2; agitation reduces particle size further and transfers the sample onto the adhesive Kapton polymer film in the vial cap. |
| **2 min 26 sec** | **Step 5:** the YuMi robot inverts the samples again and transfers them to the X-ray diffraction plate. |
| **2 min 51 sec** | **Steps 6 & 7:** The YuMi robot unscrews each sample-loaded vial cap, inverts it, and places it back into the PXRD plate. |
| **4 min 19 sec** | **Step 8:** KUKA robot collects PXRD plate containing the 8 samples. |
| **4 min 32 sec** | **Step 9:** KUKA robot transports the sample plate to the powder X-ray diffractometer. |
| **4 min 51 sec** | **Step 10:** KUKA robot opens the doors of the X-ray diffractometer. |
| **5 min 3 sec** | **Step 11:** KUKA robot transfers the X-ray plate into the diffractometer. |
| **5 min 26 sec** | **Step 12:** KUKA closes the doors of the X-ray diffractometer and diffraction data are collected. |
| **5 min 51 sec** | KUKA robot opens doors and retrieves the X-ray plate from the diffractometer. |

**Video 2 (55 min 29 sec, recorded at 4K resolution):** Video showing full workflow run including sample processing, transport, and PXRD data acquisition with no acceleration and no captions. Note that neither of these Videos shows the initial sample preparation step (liquid dispensing and evaporation) that is carried out by the Chemspeed platform (Figures S1 & S2) prior to grinding, sample mounting and PXRD analysis; both videos start at the point where crystals have already been grown. To introduce multiple camera angles, we recorded video footage from more than one run and merged them to create the videos shown here. The timings do not vary significantly between runs. Note also that the PXRD acquisition time is artificially short for the purposes of video capture; the actual acquisition time to produce the data shown in the manuscript was approximately one hour per sample.

YouTube link: https://youtu.be/cKAELaBfqEI

Original video file (downloadable):

https://www.dropbox.com/s/x2xoej9cceovczk/Amy_WF_full_54%27_29"_no_captions.mp4?dl=0

## 2. Experimental Methods

**PXRD.** Diffraction data were collected using a Panalytical X'Pert Pro diffractometer that was controlled using a Panalytical-developed software package comprising X'Pert Data Collector and X'pert Operator Interface. The X'Pert Data Collector directly communicates with the instrument and controls the instrumental parameters, such as opening and closing the shutter for the X-ray tube, the height and position of the stage and assigning the correct geometric setup. Datasets were collected in transmission mode over the range 4 to 40 in two theta in approximately 0.013 degree steps over 60 minutes. Le Bail refinement (*1*) for both the robotic and manual data was done (by hand) using TOPAS academic software (*2*), both for benzimidazole (*3*) and for ROY (*4*). All patterns were fitted in the range $10 \leq 2\theta \leq 40°$. Backgrounds were modelled using a Chebyshev polynomial with 10 coefficients and the peak shapes were described using a modified Thompson-Cox-Hastings pseudo-Voigt profile, with a single parameter used to model asymmetry due to axial beam divergence.

**KUKA KMR iiwa mobile robot manipulator:** This KUKA mobile robot comprises two connected robots; a robotic arm mounted on top of mobile base. Together, these units were controlled using the KUKA software framework, SunriseOS3, which features the arm manipulations and a complete navigation package for the base. The programs within the software framework are written in the Java programming language. Positions of the joints were saved as coordinates in frames, which can be called within the software. The position of the robot base is determined by x and y coordinates (that is, its position on the map, see Figure 2, main text) and the angle *θ* (its orientation). Transitions between saved frames were either linear (LIN), or the calculated most efficient path between the two points (PTP). The KUKA arm has seven degrees of freedom and a maximum payload of 14 kg, which is more than adequate for the tasks described here. The mobile base has a maximum speed of 3 km h$^{-1}$ and can carry payloads of up to 200 kg. It can navigate a laboratory setting by using 2D laser scanner, which produce the map of the laboratory that is shown in Figure 2 (main text). To navigate the laboratory, several user defined positions or 'way points' were created (circled numerals in Fig. 2). These locations were taught to the robot by saving their coordinates to the base. This allows easier path planning and control of the mobile base. The robot also has a range of safety features, including collision detection for both the arm and the base. For the arm, this is activated when a force of 30 N or greater is detected and this results in the arm stopping immediately. For the base, the laser scanners control how close a user can get to the robot (this distance varies with the robot speed); below this distance threshold there is an emergency stop. This emergency stopping also applies if the map changes for the robot and there is an unexpected object in its path.

**IRB 14000 YuMi dual-arm robot:** This is a dual-armed robot supplied by ABB Robotics. Each independent arm has seven axes and can accommodate interchangeable grippers. This makes this robot suitable for a wide range of applications, including assembly, pick and place operations, and packaging. The robot was fitted with ABB Robotics' SmartGripper as an end effector on each arm. Each end effector has two flat fingers for grasping. Here, neoprene adhesive tape was added to each finger, to assist with gripping and grasping of different object shapes and different materials. The YuMi robot was controlled using the ABB RobotStudio software, which is a manufacturer specific text-based programming language (see Figure S11 for an example of this code). The programming can be done through a tablet, which allows the operator to teach the robot new tasks by guiding its arms through the desired movements. As for the KUKA cobot, this robot has safety features and stops if an unexpected force is encountered, although the force-generating capability of the YuMi robot is much lower than for the KUKA robot.

**Chemspeed FLEX LIQUIDOSE liquid dispensing platform:** A Chemspeed FLEX LIQUIDOSE platform was used for dispensing accurate volumes of stock solutions into glass sample vials prior to crystallization, and for screw-capping the sample vials once the crystallization solvent had evaporated. The platform was controlled using Chemspeed's AutoSuite software package. The automated door on the front of this platform slides vertically, up and down, and can be controlled in the AutoSuite package, thus allowing the KUKA robot to access the samples unattended. The XYZ gantry robot in the platform moves in three directions allowing access to the whole platform; it has two interchangeable heads—the needle head and the screw capper tool. Changing between the heads is automated within the platform. The deck layout can be changed to accommodate custom designs for rack holders depending on the user's requirements (see Figure S2, below). The whole platform is connected to an extraction outlet to ensure safety when dispensing organic solvents.

**Grinding station 1:** The station reduces the particle size of the crystals using a Telfon-coated magnetic stirrer that is in each sample vial. The station comprises an IKA Digital Hotplate with a custom-designed 8-position 3D printed sample vial holder.

**Grinding station 2:** This station reduces the particle size further by shaking (IKA Vibrax Shaker Plate) and, more importantly, transfers the powder onto the adhesive Kapton film in the vial lid. The samples were inverted by the YuMi robot prior to shaking to facilitate this. The robot adds a lid to the samples prior to shaking (2 min 14 sec in **Video 1**) to retain the sample vials.

## 3. Crystal Structure Prediction Methods

Candidate crystal structures of ROY were taken from previously reported data (*5*).

Candidate crystal structures of benzimidazole were generated via Crystal Structure Prediction (CSP) with the following procedure.

The gas-phase molecular structure of benzimidazole was optimised at the PBE0/6-311G** level of theory using Gaussian 09 (*7*). The molecular structure was kept rigid during the CSP structure generation process.

Distributed, atom-centred multipoles up to hexadecapole were derived from the PBE0/6-311G** electron density by a distributed multipole analysis and partial charges were fitted to the multipoles (*8–10*). DMACRYS (*11*) was used with an anisotropic atom-atom force-field energy model for all lattice energy minimisations. Lattice energies were calculated using the FIT (*7*) force field and molecular charge densities were computed from a distributed multipole analysis (DMA) (*8*).

A quasi-random search of the crystal packing space was conducted in the most commonly observed space groups (SG) for organic molecular crystals using the Global Lattice Energy Explorer (GLEE) (*14*) software. The space groups analysed for Z'=1 and Z'=2 were:

| Z'=1 (non-chiral) | | | Z'=2 | | |
|---|---|---|---|---|---|
| SG number | SG name | # structures | SG number | SG name | # structures |
| 14 | P 1 2$_1$ / c 1 | 20000 | 2 | P -1 | 40000 |
| 19 | P 2$_1$ 2$_1$ 2$_1$ | 10000 | 14 | P 1 2$_1$ / c 1 | 100000 |
| 2 | P -1 | 10000 | 4 | P 1 2$_1$ 1 | 40000 |
| 4 | P 1 21 1 | 10000 | 19 | P 2$_1$ 2$_1$ 2 1 | 40000 |
| 61 | P b c a | 10000 | 1 | P 1 | 20000 |
| 15 | C 1 2 / c 1 | 20000 | 29 | P c a 2$_1$ | 10000 |
| 33 | P n a 2$_1$ | 10000 | 33 | P n a 2$_1$ | 10000 |
| 9 | C 1 c 1 | 10000 | 15 | C 1 2 / c 1 | 10000 |
| 29 | P c a 2$_1$ | 10000 | 61 | P b c a | 10000 |
| 5 | C 1 2 1 | 10000 | 5 | C 1 2 1 | 10000 |
| 1 | P 1 | 10000 | 9 | C 1 c 1 | 10000 |
| 60 | P b c n | 10000 | 7 | P 1 c 1 | 10000 |
| 7 | P 1 c 1 | 10000 | 18 | P 2$_1$ 2$_1$ 2 | 10000 |
| 18 | P 2$_1$ 2$_1$ 2 | 10000 | | | |
| 96 | P 4$_3$ 2$_1$ 2 & P 4$_1$ 2$_1$ 2 | 10000 | | | |
| 76 | P 4$_1$ & P 4$_3$ | 10000 | | | |
| 145 | P 3$_2$ & P 3$_1$ | 10000 | | | |
| 43 | F d d 2 | 10000 | | | |
| 56 | P c c n | 10000 | | | |
| 13 | P 1 2 / c 1 | 10000 | | | |
| 169 | P 6$_1$ & P 6$_5$ | 10000 | | | |
| 88 | I 4$_1$ / a | 10000 | | | |
| 148 | R -3 | 10000 | | | |
| 20 | C 2 2 2$_1$ | 10000 | | | |
| 86 | P 4$_2$ / n | 10000 | | | |
| 154 | P 3$_2$ 2 1 & P 3$_1$ 2 1 | 10000 | | | |

A total of 600,000 valid structures (280,000 and 320,000 for Z'=1 and Z'=2) were lattice energy minimised using the software package PMIN (*15*), followed by DMACRYS (*10*).

Duplicates were removed from the obtained set of structures. Initially, structures were clustered within each space group by comparison of simulated PXRD patterns generated by PLATON (*16*), obtaining 49,680 (3096 and 46,584 for Z'=1 and Z'=2) structures. Then, an additional clustering using the COMPACK algorithm (*17*) was performed only on the structures within a window of 12 kJ mol$^{-1}$ above the global minimum energy structure (1042, of which 98 and 944 for Z'=1 and Z'=2). The final set contained 961 structures (76 and 885 for Z'=1 and Z'=2).

The resulting set of 961 structures was re-optimised by plane-wave-based periodic DFT (pDFT) using the VASP package (*18,19*). The re-optimisation was performed following a three-step procedure that has been observed to accelerate the convergence of pDFT optimisations for organic crystal structures (*20*). In the first step, only atomic positions are optimised, in the second, both atomic positions and unit-cell parameters are optimised, and in the third, a final single-point calculation is performed. All VASP calculations were performed using the PBE exchange-correlation function with the Becke-Johnson-damped Grimme dispersion (GD3BJ) correction (*21,22*).

The reoptimized 961 structures were clustered using the COMPACK algorithm (17), obtaining a final set of 902 structures.

PXRD comparison was performed using the variable-cell powder difference (VC-PWDF) method (*23*). The experimental PXRD patterns were background-corrected using DASH (*24*). The 2$\theta$ range was truncated at 10° (to avoid the peaks from the adhesive Kapton tape; see Figure 3A and Figure 4A,B,D in main text), and at 40° and 35° for benzimidazole and ROY, respectively.

## 3. Supporting Figures (x 14) & Tables (x 4)

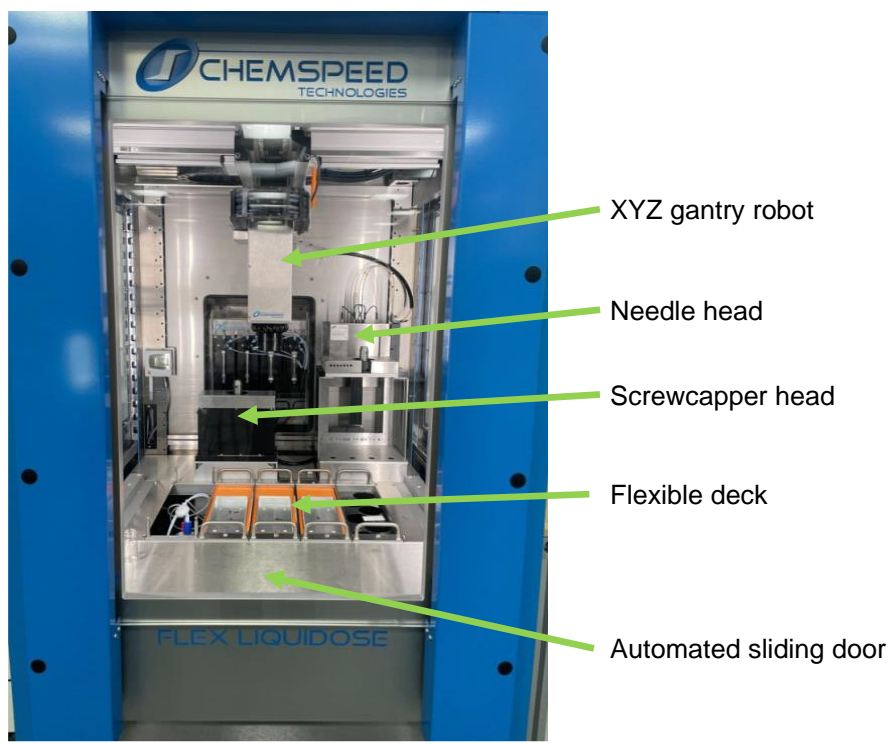

**Figure S1**: Chemspeed FLEX LIQUIDOSE platform, highlighting the vertical automated sliding door, the interchangeable needle and screw-capper heads, the flexible deck layout, and the XYZ gantry robot for sample transport.

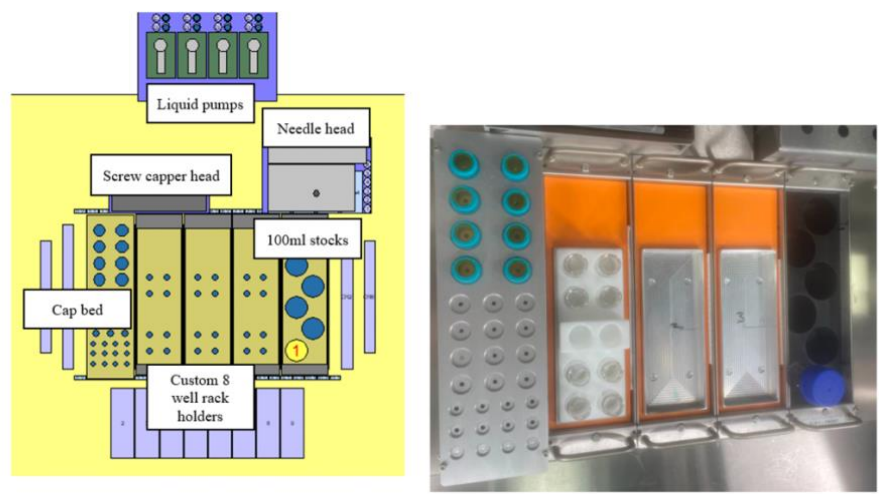

**Figure S2**: Left: Screenshot of the deck layout of the workflow, taken from the AutoSuite software. Right: Photograph of the equivalent deck layout in the Chemspeed FLEX LIQUIDOSE platform, with one occupied and two empty 8-vial racks. The Kapton-film vial caps are preloaded in the rack on the left.

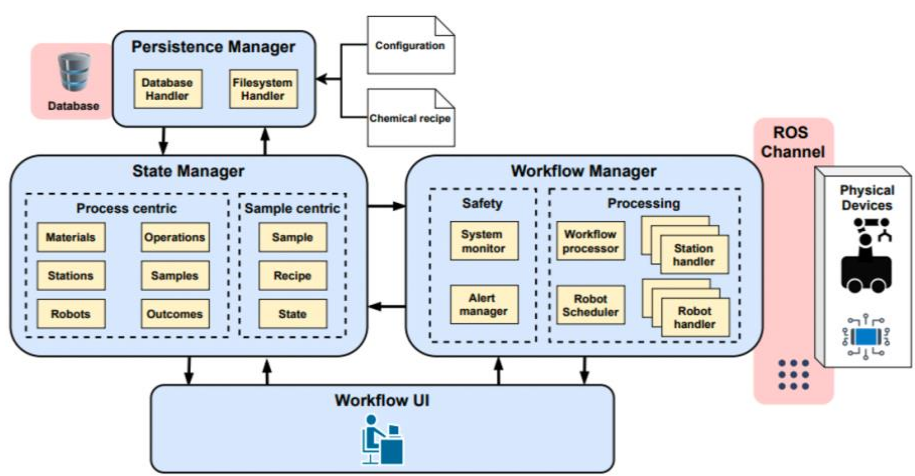

**Figure S3**: Block diagram summary of ARChemist software used to orchestrate the workflow (ref. 29, main text). The *State Manager* is responsible for representing the chemistry experiment state. The *Persistence Manager* allows the storage and retrieval of this state from a database and is responsible for parsing the input files. The *Workflow Manager* processes the experiment samples and assigns them to their respective robots and stations. The architecture uses ROS as a communication layer to interact with the physical robots and lab equipment. The *Workflow User Interface* allows the scientist to interact with the system and provide their input.

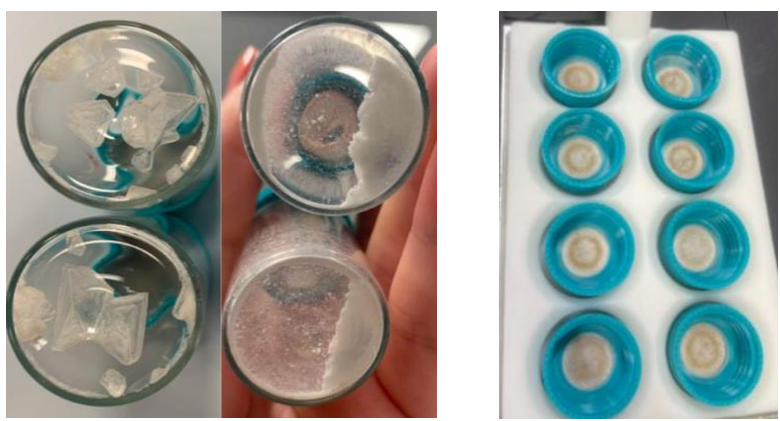

**Figure S4:** Left: Photograph comparing typical benzimidazole crystals as grown from methanol by evaporation in a glass vial, which are strongly adhered to the vial wall. Center: The same samples after automated grinding, which involved two minutes stirring and two minutes shaking with a standard magnetic stirrer in each vial (Steps 3 & 4 in Figure 1, main text). Right: Photograph showing the PXRD plate after robotic sample preparation of the benzimidazole crystals. The plate contains the 8 caps, each with an adhesive Kapton tape film (Figure S5). As described in Figure 1 and main text (see also Video 1), the robotic workflow has reduced the particle size of the large benzimidazole crystals and grinding station 2 (see Steps 4 & 5 in Figure 1, main text) has coated each film with a light dusting of the powdered benzimidazole material, which adheres to the Kapton tape.

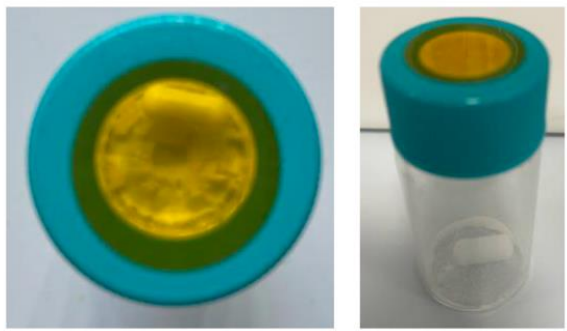

**Figure S5:** Photograph showing glass sample vials with a Kapton film covering the septum vial cap. The adhesive side of the Kapton film faces into the sample vial.

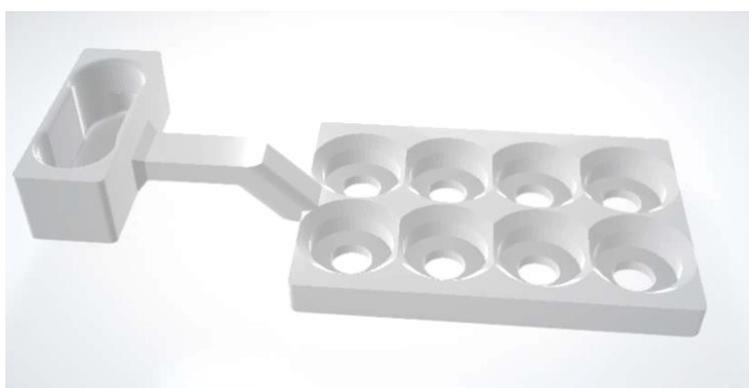

**Figure S6**: Design for the final prototype of the PXRD sample plate, which includes a custom handle that is complementary to the robot end effectors (see *e.g.*, manipulations in **Video 1**, 4 min 20 sec and 5 min 12 sec). The same design for robot grasping was also used for the sample transport plate (see 0 min 13 sec in **Video 1**). The eight sample holes are chamfered to allow easy loading of the sample caps.

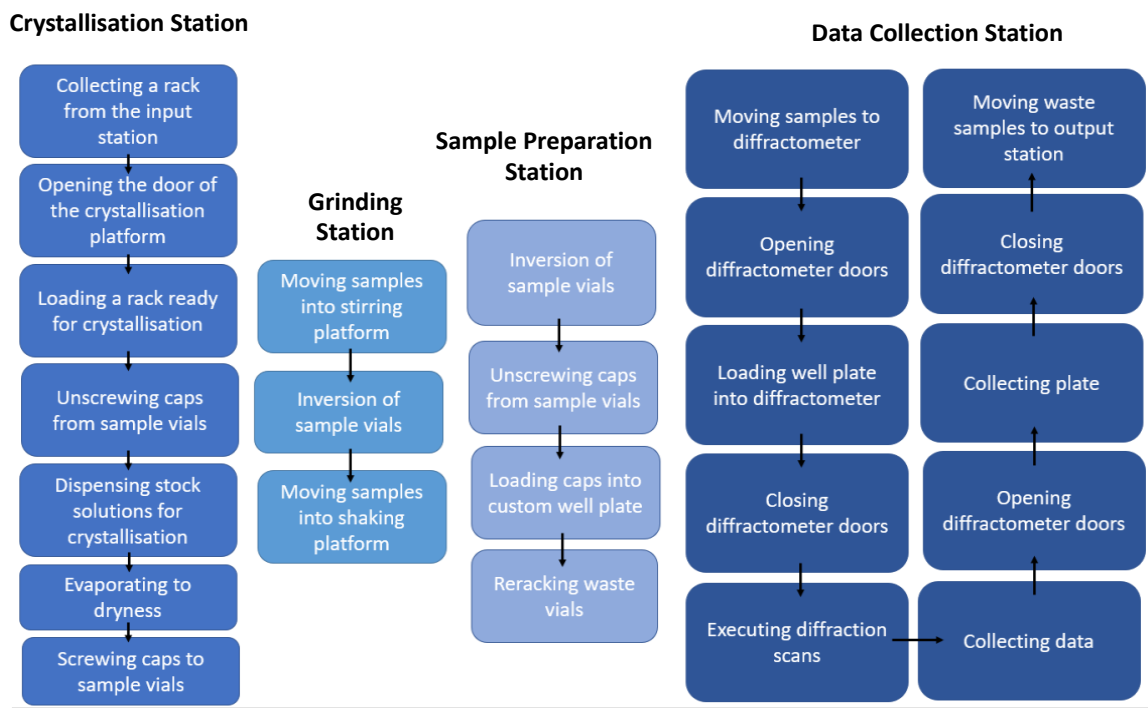

**Figure S7**: Flow diagram summarizing the various steps in the autonomous workflow.

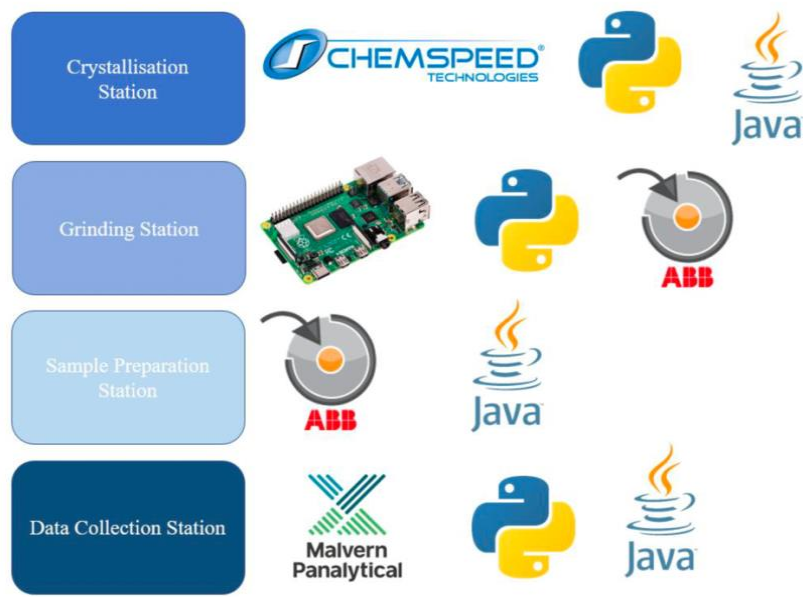

**Figure S8**: Overview of the control protocols and programming languages used for each station in the workflow. The images depict Chemspeed Autosuite GUI, Python programming language, Java programming language, Raspberry Pi microcontroller, ABB RobotStudio GUI and Malvern Panalytical GUI.

```
[ShakerPlateStation_22]: Current state changed to StationState.WAITING_ON_ROBOT
[WorkflowManager]: KukaLBRTask with task: UnloadPXRDRackYumiStation, params: ['True'] @(30, 1) is added to robot sch
eduling queue.
[KukaLBRIIWA_1]: Job (KukaLBRTask with task: UnloadPXRDRackYumiStation, params: ['True'] @(30, 1)) is assigned.
[KukaLBRIIWA_1]: Current state changed to RobotState.OP_ASSIGNED
[KukaLBRIIWA_1]: Job (KukaLBRTask with task: UnloadPXRDRackYumiStation, params: ['True'] @(30, 1)) is retrieved.
[KukaLBRIIWA_1]: Current state changed to RobotState.IDLE
[WorkflowManager]: KukaLBRIIWA_1 finished executing job KukaLBRTask with task: UnloadPXRDRackYumiStation, params: ['
True'] @(30, 1)
[WorkflowManager]: Notifying ShakerPlateStation_22
[ShakerPlateStation_22]: Robot job request is fulfilled.
[ShakerPlateStation_22]: Current state changed to StationState.PROCESSING
[ShakerPlateStation_22]: Robot job request for (KukaLBRTask with task: UnloadEightWRackYumiStation, params: ['False'
] @(30, 1)) is retrieved.
[ShakerPlateStation_22]: Current state changed to StationState.WAITING_ON_ROBOT
[WorkflowManager]: KukaLBRTask with task: UnloadEightWRackYumiStation, params: ['False'] @(30, 1) is added to robot
scheduling queue.
[KukaLBRIIWA_1]: Job (KukaLBRTask with task: UnloadEightWRackYumiStation, params: ['False'] @(30, 1)) is assigned.
[KukaLBRIIWA_1]: Current state changed to RobotState.OP_ASSIGNED
[KukaLBRIIWA_1]: Job (KukaLBRTask with task: UnloadEightWRackYumiStation, params: ['False'] @(30, 1)) is retrieved.
[KukaLBRIIWA_1]: Current state changed to RobotState.IDLE
[WorkflowManager]: KukaLBRIIWA_1 finished executing job KukaLBRTask with task: UnloadEightWRackYumiStation, params:
['False'] @(30, 1)
[WorkflowManager]: Notifying ShakerPlateStation_22
[ShakerPlateStation_22]: Robot job request is fulfilled.
[ShakerPlateStation_22]: Current state changed to StationState.PROCESSING
```

```
[KukaLBRIIWA_1]: Job (KukaLBRTask with task: LoadPXRDRackYumiStation, params: ['False'] @(30, 1) is
complete.
[KukaLBRIIWA_1]: Current state changed to RobotState.EXECUTION_COMPLETE
[IkaPlateDigital_23]: Requesting robot job (YuMiRobotTask with task: loadIKAPlate, params: [] @(30,
1))
[IKAStirPlateSm]: current state is load_stir_plate
[YuMiRobot_123]: Current state changed to RobotState.EXECUTING_OP
[INFO] [1684325354.275454]: executing loadIKAPlate
[YuMiRobot_123]: Job (YuMiRobotTask with task: loadIKAPlate, params: [] @(30, 1) is complete.
[YuMiRobot_123]: Current state changed to RobotState.EXECUTION_COMPLETE
[IkaPlateDigital_23]: Current state changed to StationState.OP_ASSIGNED
[IkaPlateDigital_23]: Requesting station job (<archemist.stations.ika_digital_plate_station.state.IK
AStirringOpDescriptor object at 0x7f86408e7430>)
[IKAStirPlateSm]: current state is stir
[IkaPlateDigital_23]: Current state changed to StationState.EXECUTING_OP
[INFO] [1684325396.002056]: executing stirring operation
[IkaPlateDigital_23]: Station op is complete.
[IkaPlateDigital_23]: Current state changed to StationState.PROCESSING
```

**Figure S9**: Two screenshots showing examples of the command line interface where the ARChemist server is launched (upper) and command line interface for the Handler's execution logs (lower). See Figure S3 for the general ARChemist overview.

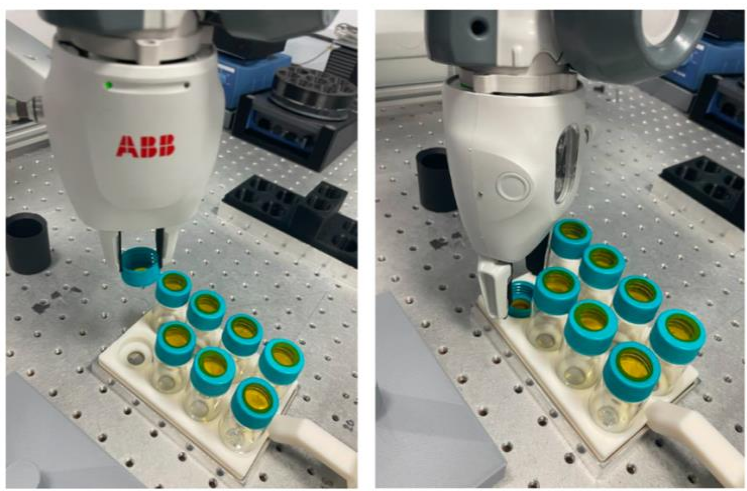

**Figure S10**: Photograph showing ABB YuMi robot placing a sample cap into the PXRD sample holder (Figure S3. In this test the sample vials are empty; when loaded with sample, the Kapton film becomes opaque (see *e.g.*, **Video 1**, 2 min 40 sec *et seq.*).

```
! unscrew
    FOR i FROM 1 TO 8 DO
        g_GripIn \holdForce:=10;
        MoveL RelTool (CRobT(\Tool:=GripperR),0,0,-1,\Rz:=-95), v200, z50,GripperR;
        WaitRob \ZeroSpeed;
        g_GripOut \holdForce:=10;
        MoveL RelTool (CRobT(\Tool:=GripperR),0,0,0,\Rz:=95), v200, z50,GripperR;
        WaitRob \ZeroSpeed;
    ENDFOR
```

**Figure S11**: Code snippet from ABB RobotStudio software used to unscrew vial caps with the YuMi robot (see Step 6 in Figure 1, main text, also **Video 1**, 2 min 49 sec *et seq*.), illustrating the intuitive programming interface that is available to construct workflows that contain multiple operations.

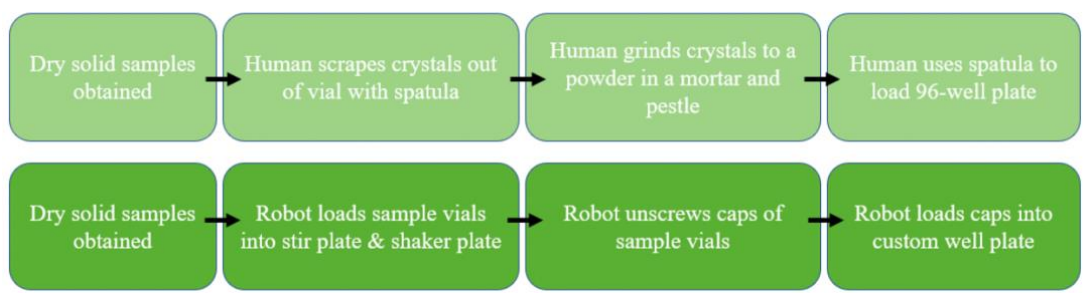

**Figure S12**: Schemes comparing typical human workflow (top) and autonomous robot workflow (bottom) for PXRD sample preparation.

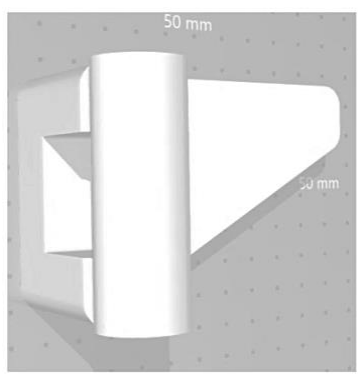

**Figure S13**: CAD file for handle modifications for the X-ray diffractometer doors; the triangular insert fits into a recess in the diffractometer doors, and this was the only physical modification made to the X-ray diffractometer, allowing the robot to open and close these doors reliably (**Video 1**, 4 min 52 sec *et seq*.) using the same end effector as used to manipulate the sample holders and PXRD platre. That the KUKA robot was, in fact, able to operate the diffractometer doors without any modification by using the existing recessed handles, but these simple 3D printed handle inserts made that process smoother and more robust.

**Table S1:** Summary of the unit cell parameters for benzimidazole obtained by Le Bail refinement (*1*) of the robotic and manual data using TOPAS academic (*2*), compared with the reported CSD (BZDMAZ01) unit cell parameters (*3*). Both PXRD datasets were collected in transmission mode over the range 4 to 40 degrees in 2θ in approximately 0.013 degree steps over 60 minutes.

|  | **Robot-processed benzimidazole** | **Manual benzimidazole** | **CSD reported benzimidazole** |
|---|---|---|---|
| *a* / Å | 13.6000(5) | 13.4918(6) | 13.507(10) |
| *b* / Å | 6.8564(2) | 6.7985(6) | 6.789(5) |
| *c* / Å | 6.9905(2) | 6.9395(4) | 6.940(5) |
| Space group | $Pna2_1$ | $Pna2_1$ | $Pna2_1$ |
| $R_{wp}$ / % | 4.69 | 19.21 | N/A |
| $\chi^2$ | 3.26 | 4.85 |  |

**Table S2:** ROY samples were prepared in 20 mL glass vials at a total solution volume of 3 mL and left to evaporate to dryness in a fume cupboard.

| **Sample number** | **Concentration of ROY in acetone (mg/mL)** | **Percentage of water in the crystallisation solution (% v/v)** |
|---|---|---|
| 1 | 25 | 20 |
| 2 | 25 | 60 |
| 3 | 25 | 50 |
| 4 | 10 | 40 |
| 5 | 10 | 50 |
| 6 | 10 | 70 |
| 7 | 10 | 80 |
| 8 | 10 | 30 |

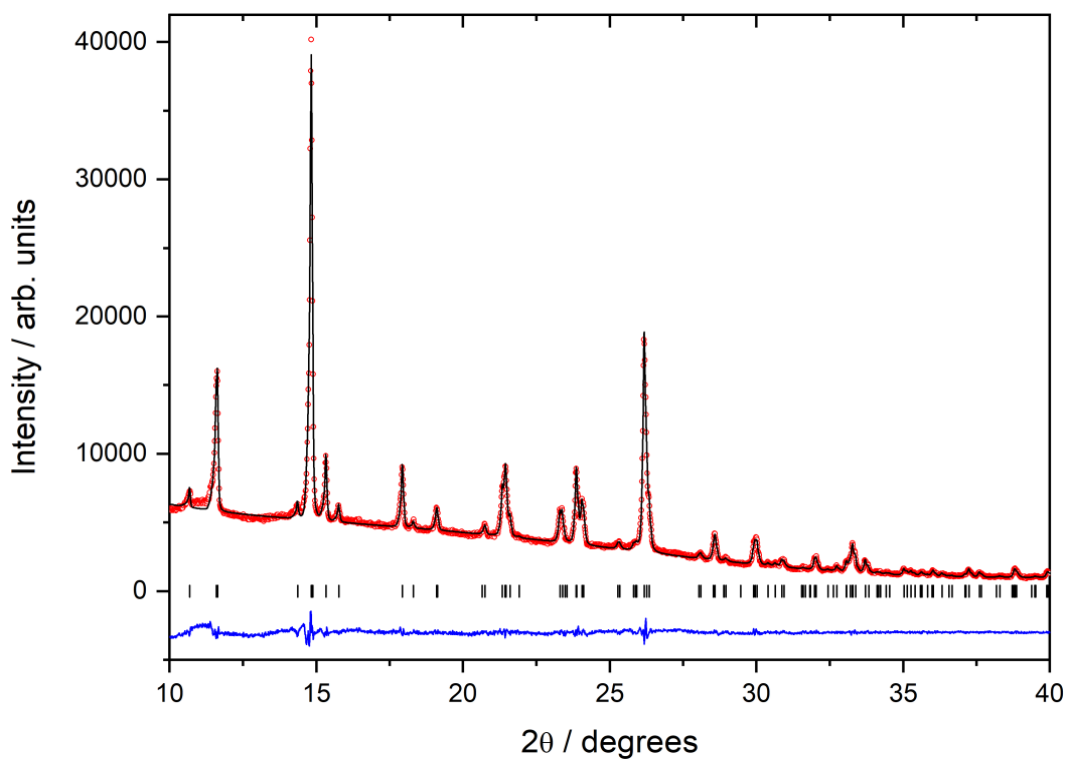

**Figure S14**: Final observed (red circles), calculated (black line) and difference (blue) profiles from Le Bail refinement (*1*) of the PXRD data obtained using TOPAS-Academic (*2*) from the robot-prepared sample 1 of ROY, as shown in Figure 4a. Tick marks indicate reflection positions.

**Table S3:** Summary of the unit cell parameters for ROY obtained by Le Bail refinement of the robotic and manual data, compared with the CSD reported (QAXMEH01) unit cell parameters (*4*).

|  | **Robotic ROY** | **Manual ROY** | **CSD reported ROY form Y** |
|---|---|---|---|
| *a* / Å | 8.6009(4) | 8.6190(8) | 8.5001(8) |
| *b* / Å | 16.5503(5) | 16.449(1) | 16.413(2) |
| *c* / Å | 8.5616(4) | 8.5790(6) | 8.5371(5) |
| *β* / ° | 91.799(3) | 91.859(6) | 91.767(7) |
| **Space group** | *P*2$_1$/*n* | *P*2$_1$/*n* | *P*2$_1$/*n* |
| *R*$_{wp}$ / % | 3.34 | 20.88 | N/A |
| $\chi^2$ | 2.06 | 10.82 |  |

**Table S4:** Powder X-ray comparison of experimental data obtained from the automated workflow with simulated powder X-ray diffraction from crystal structure prediction datasets. The parameter VC-xPWDF gives a measure of dissimilarity between the experimental pattern and the simulated pattern from the CSP structure (lower values indicate a better match). The ranking of the known structure by PXRD similarity is given, along with the relative energy of the lowest energy CSP structure that has the lowest dissimilarity (VC-xPWDF) with respect to the correct structure from the CSP set. For context, 902 CSP structures over a 20 kJ mol$^{-1}$ energy range were compared with experiment for benzimidazole. There were 264 CSP structures covering a 17.5 kJ mol$^{-1}$ range for ROY. As such, all these matches are in the top few structures in the CSP landscapes, the single worst result being for Benzimidazole 4. In other cases, the method is soemwhat robust even to phase mixtures: for example, the Y polymorph of ROY was identified as the 9$^{th}$ best match in the CSP dataset for ROY 4, even though this sample is clearly a polymorph mixture (Figure 4b, main text).

| sample | VC-xPWDF | Rank by VC-xPWDF | Lowest energy false match, kJ mol$^{-1}$ |
|---|---|---|---|
| *Benzimidazole 1* | 0.0085 | 1 | – |
| *Benzimidazole 2* | 0.0464 | 1 | – |
| *Benzimidazole 3* | 0.0710 | 6 | 5.34 |
| *Benzimidazole 4* | 0.1994 | 96 | 2.45 |
| *Benzimidazole 5* | 0.2256 | 24 | 2.56 |
| *Benzimidazole 6* | 0.1188 | 1 | – |
| *Benzimidazole 7* | 0.086 | 19 | 5.34 |
| *Benzimidazole 8* | 0.1139 | 2 | 2.71 |
| *ROY 1* | 0.0851 | 1 | – |
| *ROY 2* | 0.1683 | 27 | 0.82 |
| *ROY 3* | 0.1820 | 37 | 0.82 |
| *ROY 4* | 0.1852 | 9 | 5.69 |
| *ROY 5* | 0.2047 | 26 | 5.24 |
| *ROY 6* | 0.1306 | 4 | 6.80 |
| *ROY 7* | 0.1021 | 1 | – |
| *ROY 8* | 0.1165 | 1 | – |

## 5. Supporting References

# 6. Code and Driver Repositories

The full software and drivers for the workflow can be found in the following repositories:

**KUKA Sunrise programs:** KUKA programs from SunriseOS for the new frames for the mobile robot operations developed in this project:

https://github.com/sgalunt/Thesis_Amy_Lunt/tree/main/Appendix%201%20KUKA%20code

**ABB RobotStudio programs:** Full ABB programs from RobotStudio for the YuMi dual-arm robot operations:

https://github.com/sgalunt/Thesis_Amy_Lunt/tree/main/Appendix%202%20YuMi%20code

**ABB YuMi dual-arm robot driver:**

https://github.com/sgalunt/Thesis_Amy_Lunt/blob/main/Appendix%203%20Drivers/yumi_driver.py

**IKA stir plate driver:**

https://github.com/sgalunt/Thesis_Amy_Lunt/blob/main/Appendix%203%20Drivers/ika_serial_driver.py

**Panalytical PXRD driver:**

https://github.com/sgalunt/Thesis_Amy_Lunt/blob/main/Appendix%203%20Drivers/pxrd_driver.py

**ARChemist Process Files:** Specifically, new files developed for this workflow:

https://github.com/sgalunt/Thesis_Amy_Lunt/tree/main/Appendix%204%20ARChemist%20code/Process%20Files

**ARChemist Configuration Files:**

https://github.com/sgalunt/Thesis_Amy_Lunt/tree/main/Appendix%204%20ARChemist%20code/Configuration%20Files

**ARChemist Recipe Files:**

https://github.com/sgalunt/Thesis_Amy_Lunt/tree/main/Appendix%204%20ARChemist%20code/Recipe%20Files